\documentclass[hidelinks]{llncs}
\usepackage[utf8]{inputenc}
\usepackage{mathtools}
\usepackage{bbding}
\usepackage{flushend}
\usepackage{amssymb}
\usepackage{amsmath}
\usepackage{breqn}
\usepackage{graphicx}
\usepackage[export]{adjustbox}[2011/08/13]
\usepackage{color}
\usepackage{booktabs}
\usepackage{multirow}
\usepackage[labelfont=bf]{caption}
\usepackage{array}
\usepackage{float}
\usepackage{caption}
\usepackage{subcaption}
\usepackage{longtable}
\usepackage{bm}
\usepackage{hhline}
\usepackage[ruled]{algorithm}
\usepackage{algpseudocode}
\usepackage{eurosym}
\usepackage{hyperref}
\usepackage{url}
\usepackage{cite}
\usepackage{listings}
\usepackage{stackengine} 
\stackMath

\lstdefinestyle{mystyle}{
	frame=single,
    aboveskip=1mm,
    belowskip=1mm
   }
\usepackage{color} 
\definecolor{mygreen}{RGB}{28,172,0} 
\definecolor{mylilas}{RGB}{170,55,241}
\usepackage{stackengine} 
\stackMath
\usepackage{enumitem}

\usepackage{etoolbox}
\newcommand{\ubold}{\fontseries{b}\selectfont}
\robustify\ubold

\usepackage{multirow}
\usepackage{threeparttable}
\usepackage{siunitx}
\usepackage{multicol}

\makeatletter
\renewcommand{\@seccntformat}[1]{%
  \ifcsname prefix@#1\endcsname
    \csname prefix@#1\endcsname
  \else
    \csname the#1\endcsname\quad
  \fi}
\newcommand\prefix@subsubsection{}
\newcommand\prefix@paragraph{}
\makeatother

\begin{document}
\title{A Hybrid Approach for Aspect-Based Sentiment Analysis Using Deep Contextual Word Embeddings and Hierarchical Attention}
%
%
\author{ Maria Mihaela Tru\c{s}c\v{a}\inst{1} \inst{(}\Envelope\inst{)} \and Daan Wassenberg\inst{2} \and  \\ Flavius Frasincar \inst{2}~\orcidID{0000-0002-8031-758X} 
\and \\ Rommert Dekker\inst{2}~\orcidID{0000-0003-3823-1990}
}
%
%
\institute{Bucharest University of Economic Studies, 010374 Bucharest, Romania \and
Erasmus University Rotterdam, Burgemeester Oudlaan 50,
	3062 PA Rotterdam, the Netherlands\\
	\email{maria.trusca@csie.ase.ro},
    \email{daan.wassenberg@hotmail.com},
    \email{frasincar@ese.eur.nl},
    \email{rdekker@ese.eur.nl}}
\maketitle              
\begin{abstract}
The Web has become the main platform where people express their opinions about entities of interest and their associated aspects. Aspect-Based Sentiment Analysis (ABSA) aims to automatically compute the sentiment towards these aspects from opinionated text. In this paper we extend the state-of-the-art Hybrid Approach for Aspect-Based Sentiment Analysis (HAABSA) method in two directions. First we replace the non-contextual word embeddings with deep contextual word embeddings in order to better cope with the word semantics in a given text. Second, we use hierarchical attention by adding an extra attention layer to the HAABSA high-level representations in order to increase the method flexibility in modeling the input data. Using two standard datasets (SemEval 2015 and SemEval 2016) we show that the proposed extensions improve the accuracy of the built model for ABSA.

\end{abstract}
%
%
%

\section{Introduction} \label{sec: Introduction}

Since the evolution of the Social Web, people have benefited from the opportunity to actively interact with others sharing content from both sides. As a result, the amount of opinionated texts has risen and people had to face the problem of filtering the extra data in order to get the desired information \cite{schwartz2004paradox}. In this context, sentiment analysis turns out to be an important tool that can find sentiments or opinions at the level of a document, sentence, or aspect \cite{liu2015sentiment}. Among all levels of analysis, the most fine-grained analysis is the one orientated to aspects \cite{schouten2015survey}. The main tasks of ABSA are target extraction (TE), aspect detection (AD), and target sentiment classification (SC). Whereas, the TE task is concerned with identification of targets, i.e., attributes of the entity of interest, the aim of the AD task is to learn aspects that have a broader meaning and refer to the targets' categories. However, in this paper, we focus only on the identification of targets' sentiments (SC task) computed at the sentence level.

Deep Neural Networks (DNNs) have recently shown a great potential for sentiment classification tasks and gradually replaced rule-based approaches. While the main advantage of DNNs architectures is flexibility, rule-based classifiers imply more manual labour that confers a higher level of domain-control. The two approaches can be easily combined in a two-step method that utilises a backup classifier for all inconclusive predictions of the main classifier. One of the first two-step sentiment classification methods utilises a dictionary-based method and a Support Vector Machine (SVM) algorithm \cite{chikersal2015sentu}. Given that this method is a bit naive, we try to tackle the sentiment classification of targets using the more refined Hybrid Approach for Aspect-Based Sentiment Analysis (HAABSA) that obtains state-of-the-art results for the SC task \cite{wallaart2019hybrid}. The first step of this hybrid method employs a domain ontology \cite{schouten2018ontology} to determine the sentiments of the given targets. All the sentences for which the ontology is inconclusive input a Left-Center-Right separated neural network with Rotatory attention (LCR-Rot) \cite{zheng2018left}, as the backup model. 

In \cite{wallaart2019hybrid} two extensions of the neural network are proposed, namely Inversed LCR-Rot and Multi-hop LCR-Rot, but since the second one was shown to be the most effective, we choose it as the backup model. In this paper, we propose two extensions for HAABSA to improve the quality of the sentiment predictions. First, we replace the non-contextual GloVe word embeddings with deep contextual word embeddings, i.e., ELMo \cite{peters2018deep} and BERT \cite{devlin2018bert} in order to better consider the semantics of words context. Second, we introduce a hierarchical attention, by supplementing the current attention mechanism with a new attention layer that is able to distinguish the importance of the high-level input sentence representations. We call the new model HAABSA++. The Python source code of our extensions can be found at \url{https://github.com/mtrusca/HAABSA_PLUS_PLUS}.

The rest of the paper is organized as follows. Section \ref{Related_Works} briefly introduces the related works. Section \ref{Data_Spec} presents the details of the utilised datasets. Section \ref{Method} discusses the hybrid approach together with the extensions we propose and Sect. \ref{Evaluation} presents the experimental settings and the evaluation of our methods. Section \ref{Conclusion} gives our conclusions and suggestions for future work.

\vspace{-0.5mm}
\section{Related Works}\label{Related_Works}
Initially, ABSA’s main tasks were addressed using knowledge-based methods based on part-of-speech tagging models and lexicons \cite{kiritchenko2014nrc, wagner2014dcu}. 
Recently, machine learning including deep learning as a subset has turned out to be a more convenient solution with good rates of performance in Natural Language Processing (NLP). Whereas machine learning methods have proven to be more flexible, knowledge-based methods imply more manual labor, which makes them effective especially for in-domains sentiment classification. In \cite{yanase2016bunji} it was shown that these two approaches are in fact complementary. The sentiment polarities of aspects were learnt by applying an approach based on domain knowledge and a bidirectional recurrent neural network with attention mechanism. The research proves that there is not a winning option and while the neural network performs better for the laptop reviews of the SemEval 2015 dataset \cite{pontiki2015semeval}, the approach based on domain rules is more effective in the restaurant domain dataset of the same SemEval workshop. 

Recently, hybrid models that take advantage of both approaches in a mixed solution have been investigated in various studies. For instance, in \cite{schouten2017ontology} an SVM model was trained for target sentiment classification on an input created based on the binary presence of features identified using a domain-specific ontology. Another option to enrich the input of a neural network using domain knowledge is presented in \cite{do2018aspect}, where a self-defined sentiment lexicon is used to extend the word embeddings. Similar to our work, the neural network described in \cite{he2018effective} aims to learn context-sensitive target embeddings. Next, the attention scores are computed only for relevant words of the context indicated by a dependency parser. While the previous methods focused on integrating rule-based approaches in machine learning, in \cite{cambria2018senticnet} it is presented a different method where machine learning is used for building domain knowledge. Namely, a Long Short-Term Memory (LSTM) model with an attention mechanism is employed to create a sentiment dictionary called SenticNet 5.



Instead of integrating the two approaches in a single model, another option is to apply them sequentially \cite{chikersal2015sentu}. This option has been demonstrated to be superior to the individual approaches in \cite{schouten2018ontology}. Namely, in \cite{schouten2018ontology} an ontology developed for restaurant domain reviews is used as the first method for sentiment classification (\textit{positive} and \textit{negative}). The backup model, triggered when the ontology is inconclusive, employs a bag-of-words approach trained with a multi-class SVM associated with all three sentiment polarities (\textit{positive}, \textit{neutral}, and \textit{negative}). This work inspired \cite{mevskele2019aldona} where the SVM model is replaced with a neural network that assigns polarities to the aspects using multiple attention layers. The first one captures the relation between aspects and their left and right contexts and generates context-dependent word embeddings. The new word vectors together with sentences and aspects embeddings created using the bag-of-words approach feed the last layer of attention. 

The previous line of research is kept in \cite{wallaart2019hybrid}, where the same ontology is used together with a Multi-Hop LCR-Rot model as backup. 
Knowing the effectiveness of the two-step approach for the SC task, and considering that the method proposed in \cite{wallaart2019hybrid} achieves the best results for the SemEval 2015 and the SemEval 2016 \cite{pontiki2016semeval} datasets, we choose it as basis for our investigation on the benefits of contextual word embeddings. In addition, inspired by the hierarchical attention approach presented in \cite{yang2016hierarchical} we add to the architecture of Multi-hop LCR-Rot a new attention layer for high-level representations of the input sentence.


\vspace{-0.1mm}

\section{Datasets Specification}\label{Data_Spec}

The data used in this paper was introduced in the SemEval 2015 and 2016 contests to evaluate the ABSA task and is organised as a collection of reviews in the restaurant domain. Each review has a variable number of sentences and each sentence has one or more aspect categories. Each aspect is linked to one target that has assigned a sentiment polarity (\textit{positive}, \textit{neutral}, and \textit{negative}). Table \ref{table:table_1} lists the distribution of sentiment classes in the SemEval 2015 and SemEval 2016 datasets.
 
\begin{table}[t]
\vspace{-0.5cm}
\footnotesize
\centering
\setlength{\tabcolsep}{0pt} 
\caption{Polarity frequencies of SemEval 2015 and SemEval 2016 datasets (ABSA).}
\resizebox{\textwidth}{!}{%
\begin{tabular*}{\textwidth}{@{\extracolsep{\fill}\quad}lcccccc}
\toprule
& \multicolumn{3}{c}{SemEval 2015}
& \multicolumn{3}{c}{SemEval 2016} \\
\cmidrule{2-4}\cmidrule{5-7}
& Positive & Neutral & Negative & Positive & Neutral & Negative \\
\midrule
Train              & 72.4\% & 24.4\% & 3.2\% & 70.2\% & 3.8\% & 26.0\% \\
Test               & 53.7\% & 41.0\% & 5.3\% & 74.3\% & 4.9\% & 20.8\%\\

\bottomrule
\end{tabular*}}
\label{table:table_1}
\vspace{-0.5cm}
\end{table}

\section{Method}\label{Method}

HAABSA is a hybrid approach for aspect-based sentiment classification with two steps. First, target polarities are predicted using a domain sentiment ontology. If this rule-based method is inconclusive, a neural network is utilised as backup. Section \ref{Ontology} introduces the ontology-based rules for sentiment classification. Section \ref{Multi_Hop} gives an overview of HAABSA and presents our extensions based on various word embeddings and hierarchical attention. The new method is called HAABSA++, as a reminiscent of the base method name.


\subsection{Ontology-Based Rules}\label{Ontology}
The employed ontology is a manually designed domain specification for sentiment polarities of aspects that utilises a hierarchical structure of concepts grouped in three classes \cite{schouten2018ontology}. The \textit{SentimentValue} class groups concepts in the \textit{Positive} and \textit{Negative} subclasses, and the \textit{AspectMention} class identifies aspects related to sentiment expressions. The \textit{SentimentMention} class represents sentiment expressions. To compute the sentiment of an aspect, we utilise three rules, described below.

The first rule always assigns to an aspect the generic sentiment of its connected sentiment expression. The second rule identifies the aspect-specific sentiment expression and the sentiment is assigned only if the aspect and the linked expression belong to the same aspect category. The third rule finds the expression with a varying sentiment with respect to the connected aspect and the overall sentiment is inferred based on the pair aspect-sentiment expression. All these rules are mutually exclusive. 


The rule-based approach can identify only the \textit{positive} and \textit{negative} sentiments. By design, the neutral sentiment class is not modeled due to its ambiguous semantics. The ontology is inconclusive in two cases: (1) conflicting sentiment (predicting both \textit{positive} and \textit{negative} for a target) or (2) no hits (due to the limited coverage). In these cases a neural network is used as backup.

\subsection{Multi-Hop LCR-Rot Neural Network Design}\label{Multi_Hop}

The LSTM-ATT \cite{he2018exploiting, wang2016attention} model enhances the performance of the LSTM model with attention weighting and is a standard structure integrated by numerous sentiment classifiers. The LCR-Rot model \cite{zheng2018left} utilises this structure to detect interchangeable information between opinionated expressions and their contexts. In \cite{wallaart2019hybrid}, the LCR-Rot model is refined with repeated attention and the new classifier is called Multi-Hop LCR-Rot. In this paper, we explore the effect of different word embeddings on the Multi-Hop LCR-Rot model and propose a hierarchical attention structure to increase the model's flexibility.

The Multi-Hop LCR-Rot neural network splits each sentence into three parts: left context, target, and right context. Each of these three parts feeds three bi-directional LSTMs (bi-LSTMs). Then, a two-step rotatory attention mechanism is applied over the three hidden states associated with the bi-LSTMs (left context: $[h_1^l, ..., h_L^l]$, target: $[h_1^t, ..., h_T^t]$, and right context: $[h_1^r, ..., h_R^r]$, where $L$, $T$, and $R$ represent the length of the three input parts). At the first step, the mechanism generates new context representations using target information. Initially, an attention function $f$ is computed taking as input a parameterized product between the hidden states of the context and the target vector $r^{t_p}$ extracted using an average pooling operation. Considering for example the left context, the function $f$ is computed by: 
\begin {equation}
\label{equation1}
f(\underset{1 \times 1}{h^l_i}, r^{t_p}) = tanh(\underset{1 \times 2d}{h^{l^'}_i} \times \underset{2d \times 2d}{W_c^l} \times \underset{2d \times 1}{r^{t_p}_{}} + \underset{1 \times 1}{b^l_c}),
\end {equation}
where $W_c^l$ is a weight matrix, $b_c^l$ is a bias term, and $d$ represents the dimension of the $i$-th hidden state $h^l_i$ for $i = 1, ..., L$.

Then, the attention normalised scores $\alpha_i^l$ associated with $f$ are defined using the softmax function as follows:
\begin {equation}
\label{equation2}
\alpha_i^l = \frac{exp(f(h_i^l, r^{r_p}))}{\sum_{j=1}^L exp(f(h_j^l, r^{r_p}))}.
\end {equation}

In the end, context representations are computed using hidden states weighted by attention scores. For example, the left target2context vector is defined as:
\begin {equation}
\label{equation3}
\underset{2d \times 1}{r^l_{}} = \sum_{i = 1}^{L}\underset{1 \times 1}{\alpha^l_i} \times \underset{2d \times 1}{h_i^l}.
\end {equation}

At the second step of the rotatory attention, target representations are computed similarly, following the previous three equations. The only difference is that instead of the $r^{t_p}$ vector that stands for target information, the left and right contexts vectors ($r^l$ and $r^r$) are employed to obtain a better target representation. Taking again the left context as example, the left context2target representation $r^{t_l}$ is:
\begin {equation}
\label{equation4}
\underset{2d \times 1}{r^{t_l}_{}} = \sum_{i = 1}^{T}\underset{1 \times 1}{\alpha_i^{t_l}} \times \underset{2d \times 1}{h_i^t},
\end {equation}
where $\alpha^{t_l}$ represents the target attention scores with respect to the left context computed as above.

The right vectors, target2context and context2target ($r^r$ and $r^{t_r}$) are computed in a similar way. In a multi-hop rotatory attention mechanism, the two aforementioned steps are applied sequentially for $n$ times. In \cite{wallaart2019hybrid} the optimal $n$ value is three (the trials were executed for four scenarios: $n = \overline{1,4}$). One should note that the $r^{t_p}$ target vector computed using average pooling is used only for the first iteration of the rotatory attention. At the next iterations the vector $r^{t_p}$ is replaced with one of the vectors $r^{t_l}$ or $r^{t_r}$, depending on the considered context. At the end of the rotatory attention, all the four vectors are concatenated and feed an MLP layer for the final sentiment prediction. 

The learning process is realised using a backpropagation algorithm by minimising the cross-entropy loss function with L2 regularization. All weight matrices and biases are initialised by a uniform distribution and are updated using stochastic gradient descent with a momentum term.

\subsection{Word Embeddings}\label{Embeddings}

The first proposed extension examines the effect of deep context-dependent word embeddings on the overall performance of the neural network. Since the Multi-Hop LCR-Rot model already captures shallow context information for each target of a sentence, it is important to analyse how this architecture is possibly improved when we use deep context-sensitive word representations. Hereinafter, we give a short description of some of the most well-known contextual and non-contextual word embeddings.

\subsubsection{Non-contextual Word Embeddings. }Non-contextual word embeddings are unique for each word, regardless of its context. As a result, the polysemy of words and the varying local information are ignored. GloVe, word2vec, and fastText context-independent word embeddings are presented below.

\paragraph{GloVe. }The GloVe model generates word embeddings using word occurrences instead of language models (like word2vec), which means that the new word embeddings take into account global count statistics, instead of only the local information \cite{pennington2014glove}. The idea behind the GloVe model is to determine two word embeddings $w_i$ and $w_k$ for words $i$ and $k$, respectively, whose dot product is equal with the logarithmic value of their co-occurrence $X_{ik}$. The relation is adjusted using two biases ($b_i$ and $b_k$) for both words $i$ and $k$ as follows:
\begin {equation}
\label{equation5}
{w^T_i}{w_k} + {b_i} + {b_k} = log({X_{ik}}).
\end {equation}
The optimal word embeddings are computed using a weighted least-squares method using the cost function defined as:
\begin {equation}
\label{equation6}
J = \sum_{i=1}^{V} {\sum_{k = 1}^{V}f(X_{ik})({w^T_i}{w_k} + {b_i} + {b_k} - log({X_{ik}}))^2},
\end {equation}
where $V$ is the vocabulary size and $f(X_{ik})$ is a weighting function that has to be continuous, non-decreasing, and to generate relatively small values for large input values. The last two conditions for $f$ are necessary to prevent over-weighting of either rare or frequent co-occurrences. In this paper, we choose to use 300-dimension GloVe word embeddings trained on the Common Crawl (42 billion words) \cite{pennington2014glove}.

\paragraph{Word2vec. }The word2vec word embeddings were the first widely used word representations and since their introduction they have shown a significant improvement for many NLP tasks. The word2vec model works like a language model that facilitates generation of the more close word representations in the embedding space for words with similar context \cite{mikolov2013distributed}. The word2vec model has two variations: Continuous-Bag-Of-Words (CBOW) and Skip-Gram (SG). CBOW word embeddings represent the weights of a neural network that maximize the likelihood that words are predicted from a given context of words and SG does it the other way around. Both variations exploit the bag-of-words approach and the sequencing of words in the given or predicted context of words is irrelevant. The CBOW and SG models are trained using the following loss functions:
\begin {equation}
\label{equation7}
CBOW: J = \frac{1}{V}\sum_{t=1}^{V}\log p(w_t|w_{t-c}, \dots, w_{t-1}, w_{t+1}, \dots, w_{t+c}),
\end {equation}
\begin {equation}
\label{equation8}
SG: J = \frac{1}{V}\sum_{t=1}^{V}\sum_{i=t-c,i\neq t}^{t+c} \log p(w_i|w_t).
\end {equation}
where [-c, c] is the word context of the word $w_t$.

CBOW is considered to be faster to train than SG, but SG benefits of a better accuracy for non-frequent words \cite{ay2018evaluating}. Therefore, in the present word, both variations of word2vec are examined. The pre-trained word2vec word embeddings we use are already trained on Google News dataset (100 billion words) and their length is 300 features.

\paragraph{FastText. }The fastText model computes non-contextual word embeddings using a word2vec SG approach where the word context is represented by its n-grams \cite{bojanowski2017enriching}. As a result, out-of-vocabulary words are better handled as they can benefit from representations closer to the ones of in-vocabulary words with similar meaning in the embedding space. Given that our employed datasets are small, we utilise already computed fastText word embeddings trained on statmt.org news, UMBC webbase corpus, and Wikipedia dumps ((16 billion words). The dimensionality of word embeddings is 300.

\subsubsection{Contextual Word Embeddings. }Contextual word embeddings take into account the context of words which means that they handle better the semantics and the polysemy. Below we focus on ELMo and BERT deep contextual word embeddings.


\paragraph{ELMo. }The ELMo word embeddings capture information about the entire input sentence using multiple bidirectional LSTM (bi-LSTM) layers \cite{peters2018deep}. The main difference between the ELMo model and other language models developed on LSTM layers is that ELMo word embeddings integrate the hidden states of  all $L$ bi-LSTMs layers in a linear combination instead of utilising only the hidden states of the last layer. The ELMo model can be considered a task-specific language model that can be adjusted to different computational linguistic tasks by learning  different weights for all LSTM layers. ELMo representation of word $i$ for a given \textit{task} $ELMo_i^{task}$ is computed as follows:
\begin {equation}
\label{equation9}
ELMo_i^{task} = \gamma^{task}\sum_{j = 0}^Ls_j^{task}h_{i, j},
\end {equation}
where $h_{i, j}$ represents the concatenated hidden states of the $j$ bi-LSTM layer ($h_{i, j}$ = [$\overset{\rightarrow}{h}_{i, j}, \overset{\leftarrow}{h}_{i, j}$]), $s_j^{task}$ is its weight, and $\gamma^{task}$ scales the word embeddings accordingly to the given task.

The model we use to generate ELMo word embeddings employs two bi-LSTM layers with 512 dimension hidden state which means the size of the final word embeddings is 1024. The model is pre-trained on the 1B Word Benchmark dataset.

\paragraph{BERT. }The BERT model unlike the ELMo language model that utilises LSTM hidden states, creates contextual word representations by averaging token vectors (unique for each vocabulary word), position embeddings (vectors for word locations in the sentence), and segment embeddings (vectors of sentence indices that contains the given word). The new sequence of word embeddings is given as input to a Transformer encoder \cite{vaswani2017attention} based on the (bidirectional) self-attention. 
The Transformer encoder has $L$ blocks and each one contains a Multi-Head Attention layer followed by a fully connected layer. The output of each block feeds the input of the next one. Each Multi-Head Attention has $A$ parallel attention layers that compute the attention scores for each word with respect to the rest of the words in the sentence. The word representations associated with each Transformer block are computed by concatenating all attention-based representations. Recently, Transformers have become more common than other widely applied neural networks like Convolutional Neural Networks (CNNs) and Recurrent Neural Network (RNNs) due to their capacity to apply the parallelization (as CNNs) and to control long-term dependencies (as RNNs). 

The BERT model is pre-trained simultaneously on two tasks: Masked Language Model (MLM) and Next Sentence Prediction (NSP) using BookCorpus (800 million words) and Wikipedia dumps (2,500 million words). The first task employs a bidirectional Transformer to predict some masked words and the second task tries to learn sequence dependencies between sentences. The final loss function is computed as a sum of the task losses. In this paper, BERT word embeddings are generated using the pre-trained BERT Base model (L=12, A=12, H=768), where H stands for hidden states and represents the size of the word embeddings. The final representations of the word $i$ is computed by summing the word embeddings of the last four layers (as it was suggested in \cite{devlin2018bert}):
\begin {equation}
\label{equation10}
BERT_i = \sum_{j = 9}^{12}H_{i, j}.
\end {equation}

\subsection{Multi-Hop LCR-Rot with Hierarchical Attention}\label{Hierarchical}

The main disadvantage of Multi-Hop LCR-Rot is that the four target2context and context2target vectors are computed using only local information. Hierarchical attention alleviates this process by providing a high-level representation of the input sentence that updates each target2context and context2target vector with a relevance score computed at the sentence level. The final sentiment prediction considers the newly obtained vectors.

\begin{figure}[!b]
\vspace{-0.5cm}
\centering
\includegraphics[width=0.95\columnwidth]{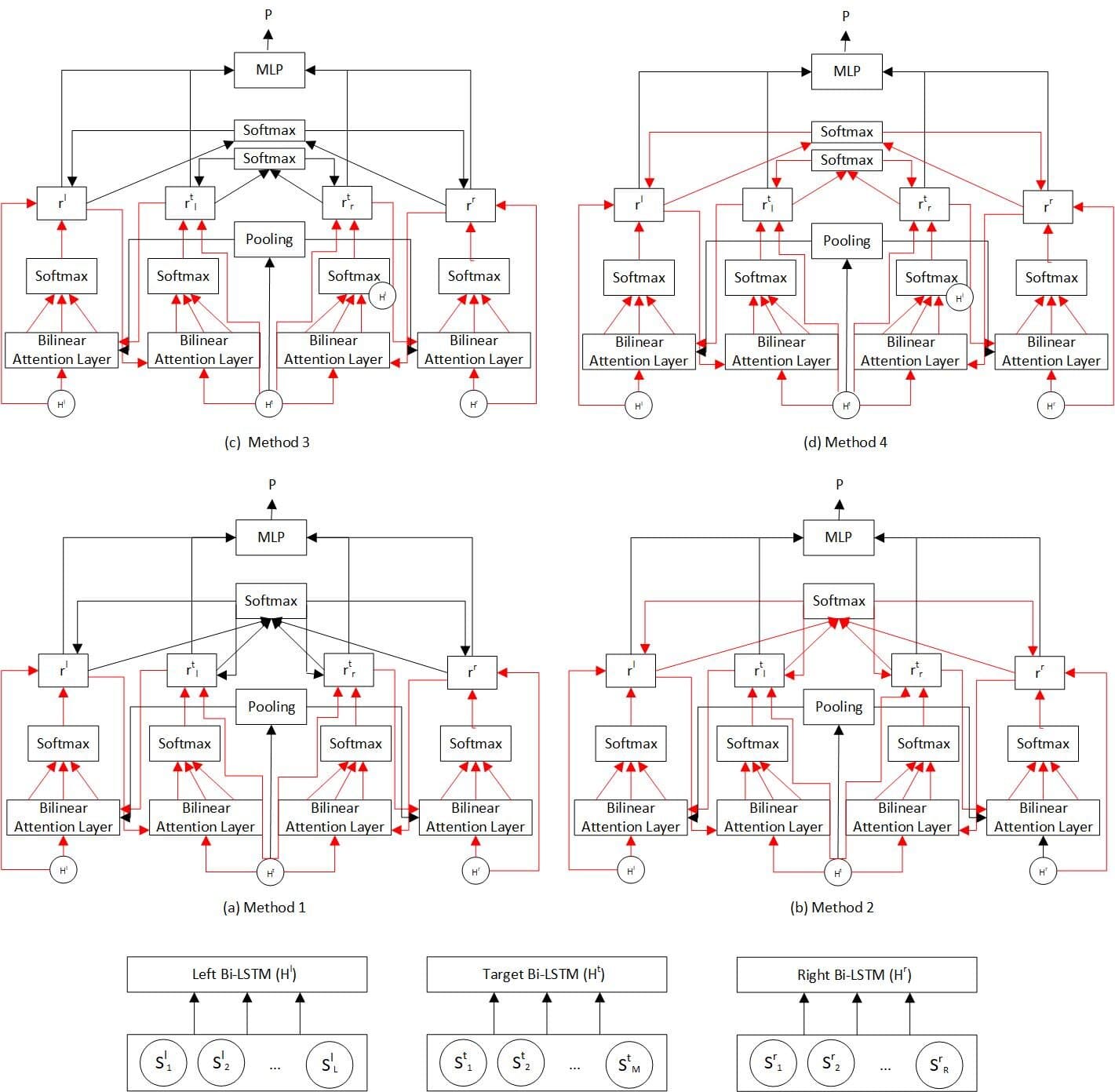}
\caption{Multi-Hop LCR-Rot with hierarchical attention} 	
\label{fig:figure_1}
\end{figure} 

First, we have to compute an attention function $f$ defined as:
\begin {equation}
\label{equation11}
f(\underset{1 \times 1}{v^i_{}}) = tanh(\underset{1 \times 2d}{v^{i^'}_{}} \times \underset{2d \times 1}{W_{ }^{ }} + \underset{1 \times 1}{b_{ }^{ }}),
\end {equation}
where $v^i$ is the representation $i$ of the input sentence ($v^i \in \{r^r, r^l, r^{t_r},r^{t_l}\}$, $i = \overline{1,4}$), $W$ is a weight matrix, and $b$ is a bias. The attention function $f$ is used to compute new attention scores $\alpha^i$ for each input $v^i$:
\begin {equation}
\label{equation12}
\alpha^i = \frac{exp(f(v^i)}{\sum_{j=1}^4 exp(f(v^j))}.
\end {equation}
The new scaled context2target or target2context vectors are:
\begin {equation}
\label{equation13}
\underset{2d \times 1}{v^i_{ }} = \underset{1 \times 1}{\alpha^i_{ }} \times \underset{2d \times 1}{v^i_{ }},
\end {equation}
	
We consider four methods to introduce hierarchical attention in the architecture of the Multi-Hop LCR-Rot model:
\begin{itemize}

    \item \textbf{Method 1}: attention weighting is applied on the final four vectors of the rotatory attention (Fig. \ref{fig:figure_1} (a)).
    \item \textbf{Method 2}: attention weighting is applied in each iteration of the rotatory attention, on the intermediate four vectors (Fig. \ref{fig:figure_1} (b)).
    \item \textbf{Method 3}: attention weighting is separately applied on the final two context and target vectors pairs of the rotatory attention (Fig. \ref{fig:figure_1} (c)).
    \item \textbf{Method 4}: attention weighting is separately applied in each iteration of the rotatory attention, on the intermediate context and target vectors pairs (Fig. \ref{fig:figure_1} (d)).
\end{itemize}

To optimise the performance of the newly proposed methods based on hierarchical attention, we have to tune again some of the model's hyperparameters like the learning rate, the momentum term, the L2 regularization term, and the dropout rate (applied to all hidden layers). The algorithm we employ for tuning is a tree-structured Parzen estimator (TPE) \cite{bergstra2011algorithms}.

\section{Evaluation}\label{Evaluation}

We compare our extensions with the baseline Multi-Hop LCR-Rot neural network, a state-of-the-art model in the SC task for both SemEval 2015 and SemEval 2016 datasets. Like \cite{wallaart2019hybrid}, our main classifier is a domain sentiment ontology. The importance of the hybrid method is pointed out in \cite{schouten2018ontology} where all the inconclusive cases of the domain sentiment ontology are assigned to the majority class of the dataset. The accuracy reported for the reference approach on the SemEval datasets is 63.3\% and 76.1\%, respectively, much lower than the accuracy of the hybrid approach.

\begin{table}[!htp]
\footnotesize
\centering
\setlength{\tabcolsep}{0pt} 
\caption{\footnotesize {Comparison of word embeddings for the Multi-Hop LCR-Rot model using accuracy. The best results are given in bold font.}}
\resizebox{\textwidth}{!}{%
\begin{tabular*}{\textwidth}{@{\extracolsep{\fill}\quad}lcccc}
\toprule
& \multicolumn{2}{c}{SemEval 2015}
& \multicolumn{2}{c}{SemEval 2016} \\
\cmidrule{2-3}\cmidrule{4-5}
& in-sample & out-of-sample & in-sample & out-of-sample \\
\midrule
\multicolumn{1}{l}{\itshape Context-independent word embeddings} \\
GloVe (HAABSA)              & \textbf{88.0}\% & 80.3\% & 89.6\% & 86.4\%  \\
CBOW              & 84.8\% & 74.6\% & 82.7\% & 83.5\%  \\
SG              & 84.7\% & 76.0\% & 85.4\% & 84.1\%  \\
FastText              & 87.4\% & 79.0\% & 87.3\% & 86.5\%  \\
\midrule
\multicolumn{1}{l}{\itshape Context-dependent word embeddings} \\
ELMo              & 85.1\% & 80.1\% & \textbf{91.1}\% & \textbf{86.7}\%  \\
BERT              & 87.9\% & \textbf{81.1}\% & 89.2\% & \textbf{86.7}\%  \\

\bottomrule
\end{tabular*}}
\label{table:table_2}
\vspace{-0.5cm}
\end{table}

The evaluation is done in terms of training and testing accuracy. Since our work is an extension of the baseline model, we re-run the Multi-Hop LCR-Rot to assure a fair comparison. First, the embedding layer is optimised by trying different word embeddings; the results thereof are shown in Table \ref{table:table_2}. Given that our base model \cite{wallaart2019hybrid} utilises the GloVe embeddings, we start presenting the results for context-independent word representation models. CBOW and SG models lead to the worst predictions and, as it is already expected, the SG model performs better than CBOW by 1.4\%-0.6\%.
The difference between the performance of the fastText and SG models is equal to three percentage points in the SemEval 2015 test dataset, which means that the fastText model is clearly an improvement of the SG model. Even if fastText outperforms the GloVe model by 0.1\% for the SemEval 2016 test dataset, given the overall performance of the GloVe model, we can conclude that it is the best context-independent word representation option. 

\begin{table}[bp]
\vspace{-0.7cm}
\centering
\footnotesize

\setlength{\tabcolsep}{0pt} 
\caption{\footnotesize {Comparison between the four methods proposed for HAABSA++ using accuracy. The best results are given in bold font.}}
\resizebox{\textwidth}{!}{%
\begin{tabular*}{\textwidth}{@{\extracolsep{\fill}\quad}lcccc}
\toprule
& \multicolumn{2}{c}{SemEval 2015}
& \multicolumn{2}{c}{SemEval 2016} \\
\cmidrule{2-3}\cmidrule{4-5}
& in-sample & out-of-sample & in-sample & out-of-sample \\
\midrule
Method 1              & 87.9\% & 81.5\% & 88.0\% & \textbf{87.1}\%  \\
Method 2              & 87.9\% & \textbf{81.7}\% & 88.7\% & 86.7\%  \\
Method 3              & 87.8\% & 81.3\% & 88.7\% & 86.7\%  \\
Method 4              & \textbf{88.0}\% & \textbf{81.7}\% & \textbf{88.9}\% & 87.0\%  \\
\bottomrule
\end{tabular*}}
\label{table:table_3}
\end{table}

\begin{table}[htp]
\footnotesize{}
\centering

\setlength{\tabcolsep}{5pt} 
\caption{\footnotesize {Comparison between HAABSA++ (Method 4) with state-of-the-art models in SC task using accuracy. SW stands for the SemEval Winner (the most effective result reported in the SemEval contest). The best results are given in bold font.}}
\resizebox{\textwidth}{!}{%
\begin{tabular}{l c l c}
\toprule
\multicolumn{2}{c}{SemEval 2015} 
& \multicolumn{2}{c}{SemEval 2016} \\
\midrule
HAABSA++ (Method 4)              & \textbf{81.7}\% & XRCE (SW) \cite{pontiki2016semeval} & \textbf{88.1}\%  \\
LSTM+SynATT+TarRep \cite{he2018effective}              & \textbf{81.7}\% & HAABSA++ (Method 4) & 87.0\%  \\
PRET+MULT \cite{he2018exploiting}             & 81.3\% & BBLSTM-SL \cite{do2018aspect} & 85.8\%  \\
BBLSTM-SL \cite{do2018aspect}            & 81.2\% & PRET+MULT \cite{he2018exploiting} & 85.6\%  \\
Sentiue (SW) \cite{pontiki2015semeval}          & 78.7\% & LSTM+SynATT+TarRep \cite{he2018effective} & 84.6\%  \\
\bottomrule
\end{tabular}}
\label{table:table_4}
\vspace{-0.7cm}
\end{table}

As regards deep contextual word embeddings, we notice that a context-sensitive approach not always leads to better results (the ELMo model outperforms the GloVe model only for the SemEval 2016 dataset). However, the BERT model seems to have the best performance, recording the same testing accuracy as the ELMo model for the SemEval 2016 datasets and exceeding the GloVe model by more than one percentage point for SemEval 2015 datasets.

The second extension we present is an adjustment of the rotatory attention to a hierarchical architecture using BERT word embeddings. Table \ref{table:table_3} shows that adding new attention layers leads to a more accurate sentiment prediction than the baseline model with BERT word embeddings listed in Table \ref{table:table_2}. Overall it is fair to consider that the best approach to tackle the hierarchical attention is the Method 4, given the small difference between the first rank and the second rank on the SemEval 2016 test dataset. 

Further on, we compare the fourth method with other similar neural networks, state-of-the-art models in SC task. The results are listed in Table \ref{table:table_4}. We do not replicate previous works and give the results as reported in papers. The best results reported in the SemEval contests are mentioned as well. While for the SemEval 2015 data, our method achieves the highest accuracy (together with the LSTM+ SynATT+TarRep \cite{he2018effective} model) for the SemEval 2016 data, it is ranked on the second position.

\begin{figure}[!b]
\vspace{-0.8cm}
\begin{center}
\includegraphics[width=1\textwidth,center]{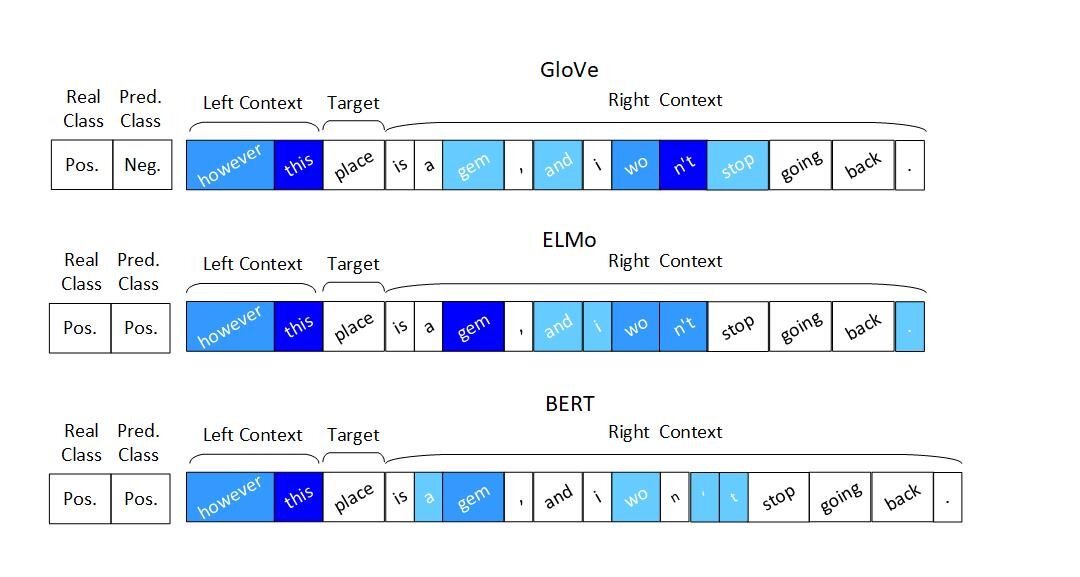}%
\end{center}
\caption{Target2Context vectors of the the Multi-Hop LCR-Rot model computed using GloVe, ELMo, and BERT word embeddings.} 	
\label{fig:figure_2}
\end{figure} 

\begin{figure}[t]
\centering
\includegraphics[width=1\textwidth,center]{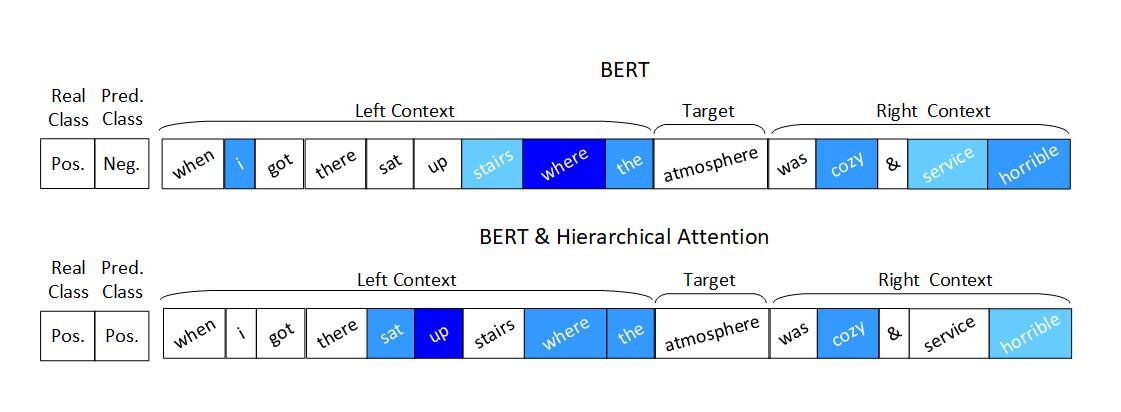}%
\caption{Target2Context vectors of the Multi-Hop LCR-Rot model with or without hierarchical attention computed using BERT word embeddings.} 	
\label{fig:figure_3}
\vspace{-0.5cm}
\end{figure} 

As we already mentioned, the Multi-Hop LCR-Rot model turns the input sentence into four vectors. Knowing that the length of the target expression is small and usually void of sentiment, we can infer that target2context vectors determine the neural network's performance to a greater extent than context2target vectors. Taking as example two sentences from the SemEval 2016 test dataset, we explore how the embedding layer and the hierarchical attention affects the predicted sentiment polarity via target2context vectors. 

Figure \ref{fig:figure_2} graphically presents attention scores associated with target2context vectors for GloVe, ELMo, and BERT word embeddings. The intensity of the blue colour shows the significance of words indicated by the attention scores. The target of the first sentence is the word ``place" and the opinionated expression (the word ``gem") indicates a positive polarity, and is located in the right context. The left context is too short and irrelevant for the target word. Only ELMo and BERT word embeddings assign the highest attention score to the opinionated word which leads to a good sentiment prediction. On the contrary, the GloVe model finds the word ``n't'' to be the most relevant for the given example, leading to a negative sentiment prediction. One should note that the BERT model has a slightly different approach to extract tokens of a sentence. This is due to the internal vocabulary used by the BERT model to guarantee the high recall on out-of-sample.

The second example explores the effect of hierarchical attention (Method 4) using BERT word embeddings. The selected sentence given in Fig. \ref{fig:figure_3} has two target expressions with different sentiment polarities. Considering the target ``atmosphere", the left context is again irrelevant while the right context contains the sentiment expression together with the second target ``service" and its opinionated expression. Even if the simple Multi-Hop LCR-Rot model without hierarchical attention assigns the highest attention scores to the words ``cozy" and ``horrible", it finds the word ``service" as relevant. As a result the sentiment prediction of the target ``atmosphere" is wrong. Differently, the neural network with hierarchical attention achieves a good prediction, considering the word ``cozy" to be the most relevant to the given target.

\section{Conclusion} \label{Conclusion}
In this work we extended the backup neural network of the state-of-the-art hybrid approach method for ABSA introduced in \cite{wallaart2019hybrid} using deep contextual word embeddings. Further on, the architecture of the model is integrated with a hierarchical structure that enforces the rotatory attention vectors to take into account high-level representations at the sentence level. Both extensions boost the testing accuracy from 80.3\% to 81.7\% for SemEval 2015 dataset and from 86.4\% to 87.0\% for SemEval 2016 dataset.


As deep learning architectures have the tendency to forget useful information from the lower layers, in future work we would like to investigate the effect of adding word embeddings to the upper layers of the architecture. Also we would like to have a better understanding of the model's inner working by applying diagnostic classification to the various layer representations.


\bibliographystyle{splncs04}
\bibliography{main}

\begin{thebibliography}{10}
\providecommand{\url}[1]{\texttt{#1}}
\providecommand{\urlprefix}{URL }
\providecommand{\doi}[1]{https://doi.org/#1}

\bibitem{ay2018evaluating}
Ay~Karaku{\c{s}}, B., Talo, M., Halla{\c{c}}, {\.I}.R., Aydin, G.: Evaluating
  deep learning models for sentiment classification. Concurrency and
  Computation: Practice and Experience  \textbf{30}(21),  e4783 (2018)

\bibitem{bergstra2011algorithms}
Bergstra, J.S., Bardenet, R., Bengio, Y., K{\'e}gl, B.: Algorithms for
  hyper-parameter optimization. In: 25th Annual Conference on Neural
  Information Processing Systems (NIPS 2011). pp. 2546--2554 (2011)

\bibitem{bojanowski2017enriching}
Bojanowski, P., Grave, E., Joulin, A., Mikolov, T.: Enriching word vectors with
  subword information. Transactions of the Association for Computational
  Linguistics  \textbf{5},  135--146 (2017)

\bibitem{cambria2018senticnet}
Cambria, E., Poria, S., Hazarika, D., Kwok, K.: Senticnet 5: Discovering
  conceptual primitives for sentiment analysis by means of context embeddings.
  In: Thirty-Second AAAI Conference on Artificial Intelligence (AAAI 2018). pp.
  1795--1802. AAAI Press (2018)

\bibitem{chikersal2015sentu}
Chikersal, P., Poria, S., Cambria, E.: {SeNTU}: Sentiment analysis of tweets by
  combining a rule-based classifier with supervised learning. In: Proceedings
  of the 9th International Workshop on Semantic Evaluation (SemEval 2015). pp.
  647--651. ACL (2015)

\bibitem{devlin2018bert}
Devlin, J., Chang, M.W., Lee, K., Toutanova, K.: {BERT}: Pre-training of deep
  bidirectional transformers for language understanding. In: 2019 Annual
  Conference of the North American Chapter of the Association for Computational
  Linguistics (NAACL-HLT 2019). pp. 4171--4186. ACL (2019)

\bibitem{do2018aspect}
Do, B.T.: Aspect-based sentiment analysis using bitmask bidirectional long
  short term memory networks. In: 31st International Florida Artificial
  Intelligence Research Society Conference (FLAIRS 2018). pp. 259--264. AAAI
  Press (2018)

\bibitem{he2018effective}
He, R., Lee, W.S., Ng, H.T., Dahlmeier, D.: Effective attention modeling for
  aspect-level sentiment classification. In: 27th International Conference on
  Computational Linguistics (COLING 2018). pp. 1121--1131. ACL (2018)

\bibitem{he2018exploiting}
He, R., Lee, W.S., Ng, H.T., Dahlmeier, D.: Exploiting document knowledge for
  aspect-level sentiment classification. arXiv preprint arXiv:1806.04346
  (2018)

\bibitem{kiritchenko2014nrc}
Kiritchenko, S., Zhu, X., Cherry, C., Mohammad, S.: {NRC-Canada-2014}:
  Detecting aspects and sentiment in customer reviews. In: 8th International
  Workshop on Semantic Evaluation (SemEval 2014). pp. 437--442. ACL (2014)

\bibitem{liu2015sentiment}
Liu, B.: Sentiment analysis: Mining opinions, sentiments, and emotions.
  Cambridge University Press (2015)

\bibitem{mevskele2019aldona}
Me{\v{s}}kel{\.e}, D., Frasincar, F.: Aldona: a hybrid solution for
  sentence-level aspect-based sentiment analysis using a lexicalised domain
  ontology and a neural attention model. In: 34th ACM Symposium on Applied
  Computing (SAC 2019). pp. 2489--2496. ACM (2019)

\bibitem{mikolov2013distributed}
Mikolov, T., Sutskever, I., Chen, K., Corrado, G.S., Dean, J.: Distributed
  representations of words and phrases and their compositionality. In: 27st
  Annual Conference on Neural Information Processing Systems (NIPS 2013). pp.
  3111--3119 (2013)

\bibitem{pennington2014glove}
Pennington, J., Socher, R., Manning, C.: Glove: Global vectors for word
  representation. In: 2014 Conference on Empirical Methods in Natural Language
  Processing (EMNLP). pp. 1532--1543. ACL (2014)

\bibitem{peters2018deep}
Peters, M.E., Neumann, M., Iyyer, M., Gardner, M., Clark, C., Lee, K.,
  Zettlemoyer, L.: Deep contextualized word representations. In: 2018
  Conference of the North American Chapter of the Association for Computational
  Linguistics-Human Language Technologies (NAACL-HLT 2018). pp. 227--2237. ACL
  (2018)

\bibitem{pontiki2016semeval}
Pontiki, M., Galanis, D., Papageorgiou, H., Androutsopoulos, I., Manandhar, S.,
  Mohammad, A.S., Al-Ayyoub, M., Zhao, Y., Qin, B., De~Clercq, O., et~al.:
  Semeval-2016 task 5: Aspect-based sentiment analysis. In: 10th International
  Workshop on Semantic Evaluation (SemEval 2016). pp. 19--30. ACL (2016)

\bibitem{pontiki2015semeval}
Pontiki, M., Galanis, D., Papageorgiou, H., Manandhar, S., Androutsopoulos, I.:
  Semeval-2015 task 12: Aspect-based sentiment analysis. In: 9th International
  Workshop on Semantic Evaluation (SemEval 2015). pp. 486--495. ACL (2015)

\bibitem{schouten2015survey}
Schouten, K., Frasincar, F.: Survey on aspect-level sentiment analysis. IEEE
  Transactions on Knowledge and Data Engineering  \textbf{28}(3),  813--830
  (2015)

\bibitem{schouten2018ontology}
Schouten, K., Frasincar, F.: Ontology-driven sentiment analysis of product and
  service aspects. In: 15th Extended Semantic Web Conference (ESWC 2018). LNCS,
  vol. 10843, pp. 608--623. Springer (2018)

\bibitem{schouten2017ontology}
Schouten, K., Frasincar, F., de~Jong, F.: Ontology-enhanced aspect-based
  sentiment analysis. In: 17th International Conference on Web Engineering
  (ICWE 2017). LNCS, vol. 10360, pp. 302--320. Springer (2017)

\bibitem{schwartz2004paradox}
Schwartz, B.: The paradox of choice: Why more is less. HarperCollins (2004)

\bibitem{vaswani2017attention}
Vaswani, A., Shazeer, N., Parmar, N., Uszkoreit, J., Jones, L., Gomez, A.N.,
  Kaiser, {\L}., Polosukhin, I.: Attention is all you need. In: 31st Annual
  Conference on Neural Information Processing Systems (NIPS 2017). pp.
  5998--6008 (2017)

\bibitem{wagner2014dcu}
Wagner, J., Arora, P., Cortes, S., Barman, U., Bogdanova, D., Foster, J.,
  Tounsi, L.: Dcu: Aspect-based polarity classification for {SemEval} task 4.
  In: International Workshop on Semantic Evaluation ({SemEval 2014}). ACL
  (2014)

\bibitem{wallaart2019hybrid}
Wallaart, O., Frasincar, F.: A hybrid approach for aspect-based sentiment
  analysis using a lexicalized domain ontology and attentional neural models.
  In: 16th Extended Semantic Web Conference (ESWC 2019). LNCS, vol. 11503, pp.
  363--378. Springer (2019)

\bibitem{wang2016attention}
Wang, Y., Huang, M., Zhu, X., Zhao, L.: Attention-based {LSTM} for aspect-level
  sentiment classification. In: Proceedings of the 2016 Conference on Empirical
  Methods in Natural Language Processing (EMNLP 2016). pp. 606--615. ACL (2016)

\bibitem{yanase2016bunji}
Yanase, T., Yanai, K., Sato, M., Miyoshi, T., Niwa, Y.: bunji at {SemEval-2016}
  task 5: Neural and syntactic models of entity-attribute relationship for
  aspect-based sentiment analysis. In: 10th International Workshop on Semantic
  Evaluation (SemEval 2016). pp. 289--295. ACL (2016)

\bibitem{yang2016hierarchical}
Yang, Z., Yang, D., Dyer, C., He, X., Smola, A., Hovy, E.: Hierarchical
  attention networks for document classification. In: 2016 Conference of the
  North American Chapter of the Association for Computational Linguistics-Human
  Language Technologies (NAACL-HLT 2018). pp. 1480--1489. ACL (2016)

\bibitem{zheng2018left}
Zheng, S., Xia, R.: Left-center-right separated neural network for aspect-based
  sentiment analysis with rotatory attention. arXiv preprint arXiv:1802.00892
  (2018)

\end{thebibliography}

\end{document}